\documentclass{article}

\usepackage[nonatbib,final]{d3s3_neurips_2024}

\usepackage[utf8]{inputenc} % allow utf-8 input
\usepackage[T1]{fontenc}    % use 8-bit T1 fonts
\usepackage{hyperref}       % hyperlinks
\usepackage{url}            % simple URL typesetting
\usepackage{booktabs}       % professional-quality tables
\usepackage{amsfonts}       % blackboard math symbols
\usepackage{nicefrac}       % compact symbols for 1/2, etc.
\usepackage{microtype}      % microtypography
\usepackage{xcolor}         % colors
\usepackage{graphicx}
\usepackage{todonotes}
\usepackage{amsmath} 
\usepackage{subcaption}
\usepackage[export]{adjustbox}
\usepackage{tabularx}
\usepackage{bm}
\usepackage{makecell} % For line breaks in table cells
\usepackage{appendix}

\title{VehicleSDF: A 3D generative model for constrained engineering design via surrogate modeling}

\author{
  Hayata Morita\\
  \And 
  Kohei Shintani\\
  \And
  Chenyang Yuan\\
  \And
  Frank Permenter\\
  \AND
  Toyota Research Institute\\
  Cambridge, MA\\
  \texttt{firstname.lastname@tri.global}\\
}

\begin{document}

\maketitle

\begin{abstract}
A main challenge in mechanical design is to efficiently explore the design space while satisfying engineering constraints.
This work explores the use of 3D generative models to explore the design space in the context of vehicle development, while estimating and enforcing engineering constraints. Specifically, we generate diverse 3D models of cars that meet a given set of geometric specifications, while also obtaining quick estimates of performance parameters such as aerodynamic drag. For this, we employ a data-driven approach (using the ShapeNet dataset) to train VehicleSDF, a DeepSDF based model that represents potential designs in a latent space witch can be decoded into a 3D model. We then train surrogate models to estimate engineering parameters from this latent space representation, enabling us to efficiently optimize latent vectors to match specifications. Our experiments show that we can generate diverse 3D models while matching the specified geometric parameters. Finally, we demonstrate that other performance parameters such as aerodynamic drag can be estimated in a differentiable pipeline.
\end{abstract}

% TLDR:
% We train VehicleSDF, a 3D generative model using surrogate models to efficiently generate and optimize diverse car designs that meet geometric and performance constraints.

\section{Introduction}

\begin{figure*}
    \centering
    \begin{subfigure}[b]{1.0\textwidth}
        \centering
        \includegraphics[valign=c,width=\textwidth]{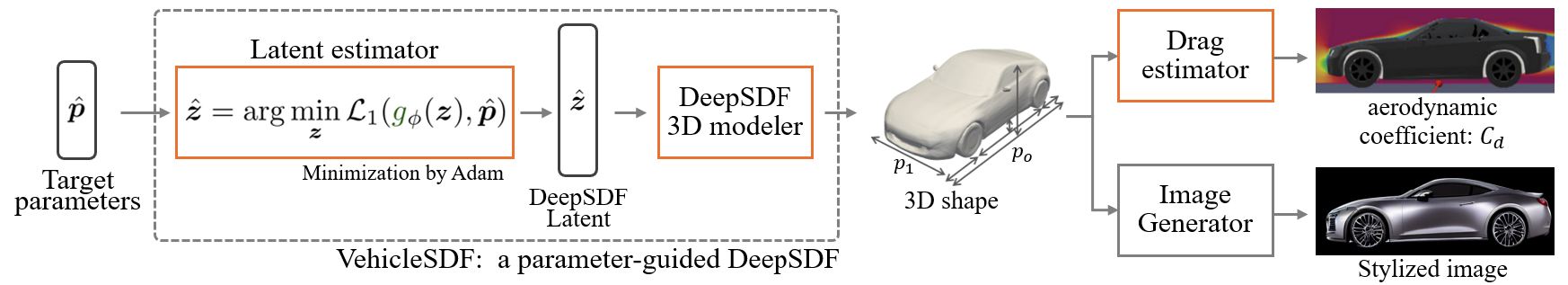}
        \caption{Proposed pipeline for vehicle design and performance estimation}
        \label{fig: Proposed pipeline for vehicle design and performance estimation}
    \end{subfigure}
    \begin{subfigure}[b]{0.47\textwidth}
        \centering
        \includegraphics[valign=c,width=\textwidth]{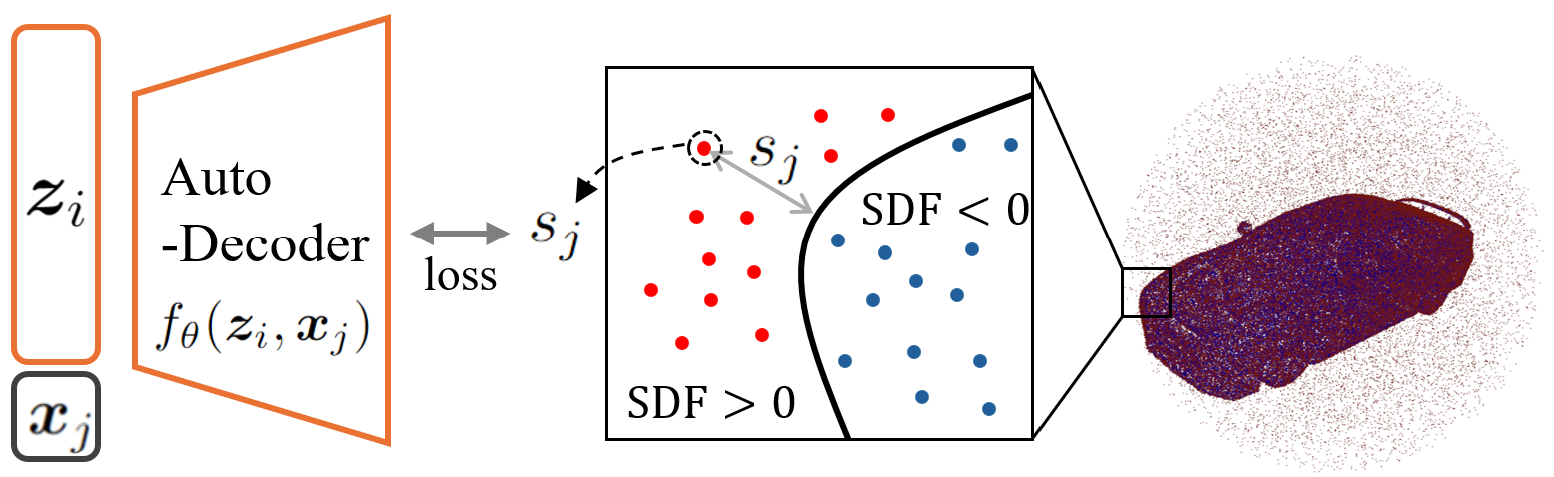}
        \caption{Training of DeepSDF using 3D shapes}
        \label{fig: Training of DeepSDF using 3D shapes}
    \end{subfigure}\;\;\;\;\begin{subfigure}[b]{0.47\textwidth}
        \centering
        \includegraphics[valign=c,width=\textwidth]{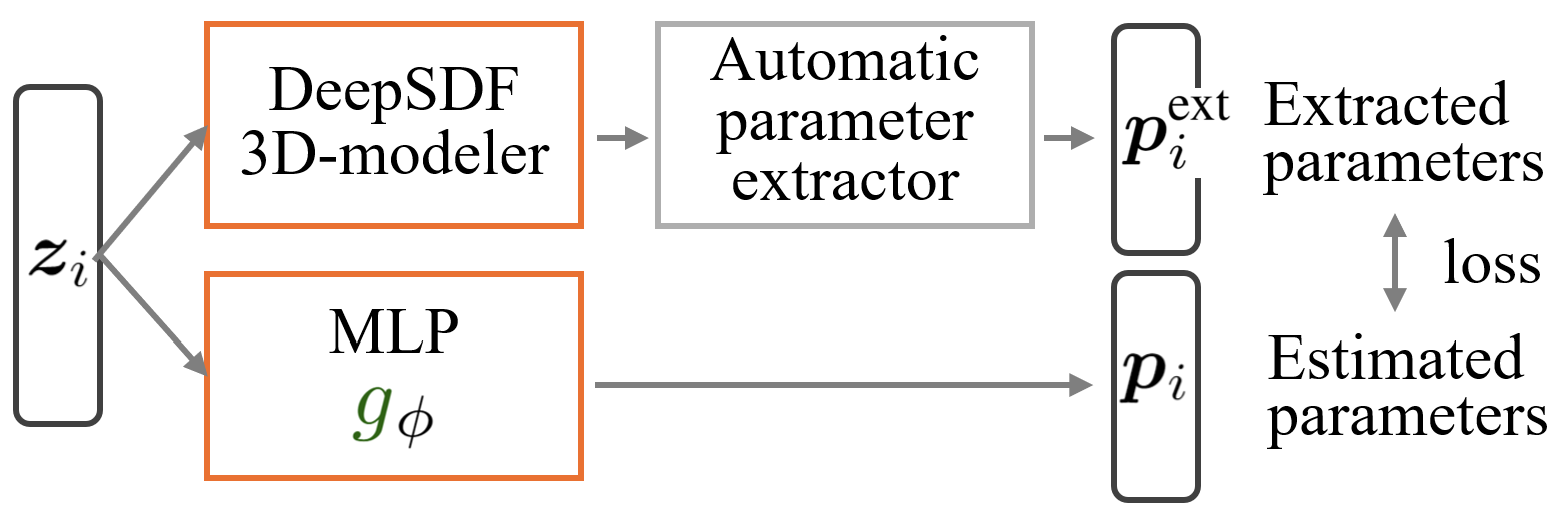}
        \caption{Training of surrogate parameter estimator}
        \label{fig: Training of surrogate parameter estimator}
    \end{subfigure}
    \caption{An illustration of our proposed pipeline and components. The latent vector \(\bm{z}_i\) in (\ref{fig: Training of DeepSDF using 3D shapes}) is initialized randomly from \(\mathcal{N}(0, \sigma^2)\) and optimized. The optimized and augmented latent vectors then are used in (\ref{fig: Training of surrogate parameter estimator}). The \textcolor{orange}{orange-outlines} mark the components we trained. }
    \label{fig: An illustration of our proposed method and components. The orange-outlines mark the components we trained.}
\end{figure*}

\begin{figure}
  \centering
  \includegraphics[width=0.75\textwidth]{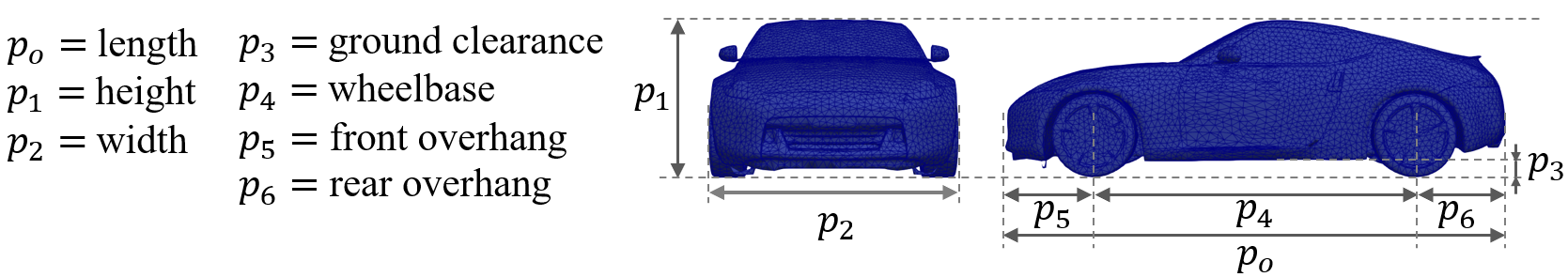}
  \caption{Vehicle geometric parameters to be specified during design}
  \label{fig: Vehicle geometric parameters to be specified during design}
\end{figure}

Recent advances in generative AI have opened new possibilities for addressing mechanical design problems while considering both mechanical performance and aesthetics at the same time~\cite{arechiga2023drag, wada2024physics}. Deep generative models have demonstrated remarkable capabilities in producing complex shapes and designs that satisfy multiple objectives simultaneously~\cite{yonekura2021data, jing2022inverse}. Despite these advancements, the integration of engineering constraints into the generative design process remains a significant challenge. This research aims to bridge this gap by proposing a 3D generative model for vehicle design that simultaneously considers geometric constraints and aesthetic styling. Our proposed method is particularly relevant in the early stages of development, where rapid iterations and evaluations are crucial. By generating diverse 3D shapes that meet specific mechanical constraints while also producing aesthetically pleasing designs, the model has the potential to streamline the design process and reduce later revisions. The proposed pipeline (Figure~\ref{fig: Proposed pipeline for vehicle design and performance estimation}) consists of three key components: %. Firstly, Secondly, Finally 

\paragraph{VehicleSDF: A 3D model generator} To generate 3D vehicle shapes matching specified geometric parameters that come from engineering constraints (Figure \ref{fig: Vehicle geometric parameters to be specified during design}), we train VehicleSDF, which is a differentiable parameter estimator combined with DeepSDF~\cite{park2019deepsdf} trained on ShapeNet~\cite{shapenet2015} dataset.

\paragraph{Drag coefficient predictor} We want to efficiently estimate engineering parameters from generated models without running expensive simulations. To demonstrate this for vehicle drag coefficient ($C_d$), we trained a surrogate model~\cite{song2023surrogate,arechiga2023drag} achieving near-instant predictions from 3D models.

\paragraph{Image generator} The final component demonstrates that we can generate photo-realistic vehicle renderings by using StableDiffusion with ControlNet~\cite{zhang2023adding} to stylize the 3D models.

By combining these components, VehicleSDF can generate diverse 3D models from vehicle geometric parameters. 
This enables designers to verify mechanical performance alongside aesthetic design during the early design phase. This integration can reduce the rework during later stage revisions.

\section{Methodology}
\label{methodology}

\subsection{VehicleSDF: A Parameter-guided DeepSDF}

We propose VehicleSDF, a parameter-guided DeepSDF for generating 3D shapes based on specified shape parameters. First, we train a DeepSDF model (Figure \ref{fig: Proposed pipeline for vehicle design and performance estimation}) on 3D vehicle data (ShapeNet) to minimize a loss function that consists of the prediction error of the SDF values at each sample point, along with an L2 norm regularization term. For further details, refer to Appendix \ref{appendix: DeepSDF}. Second, to optimize the latent vectors to ensure conformity with the target parameters, we introduce a surrogate parameter estimator (Figure \ref{fig: Training of surrogate parameter estimator}). This estimator is a multi-layer perceptron \(g_{\phi}(\bm{z}_i)=\bm{p}_i\), where \(\bm{z}_i\ \in \mathbb{R}^m\) is the latent vector, \(\bm{p}_i\ \in \mathbb{R}^n\) are the geometric parameters and \(\phi\) are the model's weights, optimized at train-time:
\begin{equation}
\textstyle
\arg\min_{\phi} \sum_{i} \mathcal{L}_\text{MSE}(g_{\phi} (\bm{z}_i), \bm{p}^{\text{ext}}_{i}) ,
\end{equation}
where \(\bm{p}^{\text{ext}}_{i}\) are parameters extracted for each shape using extraction method of Appendix \ref{appendix: Extracting parameters} and \(\mathcal{L}_1\) is the mean-squared error. Fixing $\phi$ and minimizing $\mathcal{L}_\text{MSE}(g_{\phi} (\hat{\bm{z}}), \hat{\bm{p}})$, we can find latent vectors $\hat{\bm{z}}$ matching target parameters $\hat{\bm{p}}$. 
%The extraction method refers to Appendix \ref{appendix: Extracting parameters}.

\subsection{A Surrogate Model for Drag Prediction}
\label{surrogate}

\begin{figure}
  \centering
  \includegraphics[width=1.0\textwidth]{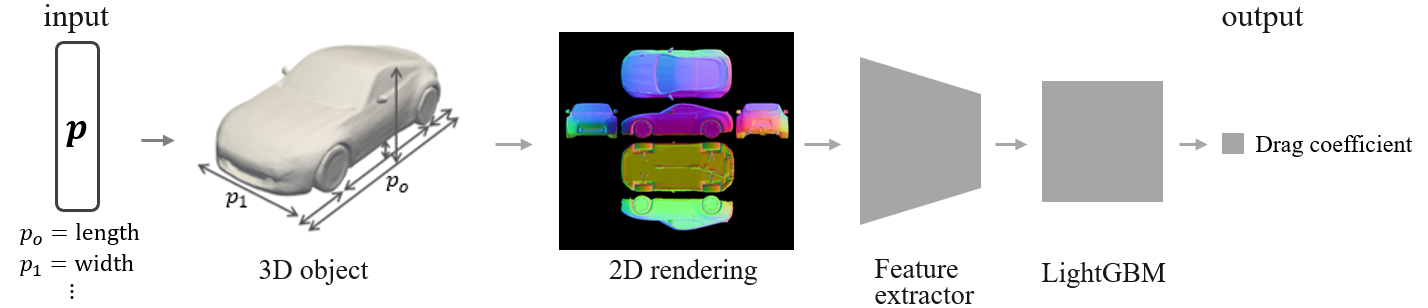}
  \caption{Drag estimation pipeline}
  \label{fig: A drag estimation}
\end{figure}

To predict \(C_d\) values, we trained a surrogate model using methods based on~\cite{song2023surrogate, arechiga2023drag} (Figure \ref{fig: A drag estimation}). Specifically, we first render the 3D object into an image that integrates normal maps from the top, bottom, left, right, front, and back views. Next, we numerically encode this image using a frozen feature extractor, and predict a \(C_d\) value from a trainable LightGBM \cite{ke2017lightgbm} layer. Our model employs established techniques in transfer learning \cite{kumar2022fine} to enhance robustness to out-of-distribution shifts. For training, we used the dataset~\cite{song2023surrogate} with the same train, validation, and test splits. The training results are discussed in Section \ref{Drag predictor}.

\subsection{Image Generation}
\label{Image-generation}
To generate stylized vehicle renderings of the generated shapes, we adopted Stable Diffusion with ControlNet \cite{zhang2023adding}. VehicleSDF generates 3D shapes, which are rendered into intermediate 2D images using either side-view normal map~\cite{vasiljevic2019diode}, depth map~\cite{ranftl2020towards}, or canny edges~\cite{canny1986computational}. These rendered images are then used as inputs to ControlNet to generate stylized images.

\begin{figure}
  \centering
  \includegraphics[width=1.0\textwidth]{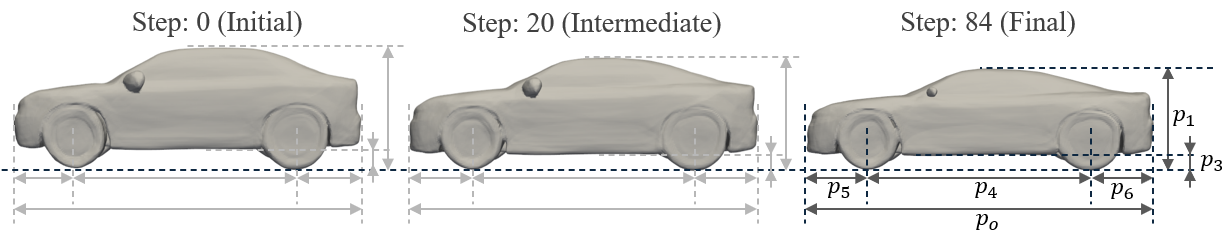}
  \caption{Optimization of 3D shape to match specified parameters}
  \label{fig: transition of 3D shape}
\end{figure}
\begin{table}
  \centering
  \caption{Comparison of target geometric parameters and during optimization}
  \begin{tabular}{lcccccccc}
  \toprule
       Parameters & $p_{0}$ & \(p_{1}\) & \(p_{2}\) & \(p_{3}\) & \(p_{4}\) & \(p_{5}\) & \(p_{6}\) & \(\text{MSE}\) \\
  \midrule
    Initial & 1.000 & 0.331 & 0.396 & 0.053 & 0.598 & 0.194 & 0.208 & $5.97\times 10^{-4}$ \\
    Intermediate & 1.000 & 0.306 & 0.425 & 0.039 & 0.599 & 0.203 & 0.199 & $1.03\times 10^{-4}$ \\
    Final & 1.000 & 0.280 & 0.431 & 0.037 & 0.600 & 0.200 & 0.200 & $2.86\times 10^{-8}$ \\
    Target  & 1.000 & 0.280  & 0.430 & 0.037 & 0.600 & 0.200 & 0.200 & - \\
  \bottomrule
  \end{tabular}
  \label{tab:vehicle_geometric}
\end{table}

\section{Experiments}
\label{Numerical example}

\paragraph{Data preparation and implementation} \label{data-preparatoin} To train VehicleSDF, we prepared the SDF sampling \(\bm{X}_i := \{(\bm{x}_j, s_j) : s_j = \text{SDF}^i(\bm{x}_j)\}\) with 55,000 spatial points and 374 watertight shapes from ShapeNet, where $\text{SDF}^i(\bm{x}_j)$ represents the SDF value of spatial point \(\bm{x}_j \in \mathbb{R}^3\) of shape $i$, calculated using~\cite{Rusu_ICRA2011_PCL}. Add reference to Appendix~\ref{appendix: DeepSDF}. We also normalized each shape to fit within a unit sphere. To train the surrogate parameter estimator, we prepared a dataset including geometric parameters and the latent vector for each shape, further described in Appendix \ref{appendix: Extracting parameters}. The parameters were normalized to unit length. To improve estimator accuracy, we also augmented the dataset using interpolated shapes generated by the trained VehicleSDF. An interpolated latent vector is defined as \( \bm{r} = (1-\alpha)\bm{a} + \alpha\bm{b} \), where \(\alpha \in (0, 1)\), \(\bm{a}\) and \(\bm{b}\) are optimized latent vector. We then augmented the training data from 374 to 10,000 shapes, after which the test MSE saturates. The final test MSE, train and test $R^2$ were $3.38\times 10^{-6}$, 0.86 and 0.85, respectively.

\paragraph{Parametric 3D model generation with VehicleSDF} We demonstrate that VehicleSDF generates 3D shapes that satisfy target geometric parameters. Figure \ref{fig: transition of 3D shape} shows the initial, intermediate, and final shapes during optimization. Here, the initial shape refers to the shape reconstructed from a randomly initialized latent vector from \(\mathcal{N}(0, 0.01^2)\). The final shape closely matches the target parameters, as seen in Table \ref{tab:vehicle_geometric}.

\begin{figure}
  \centering
  \includegraphics[width=1.0\textwidth]{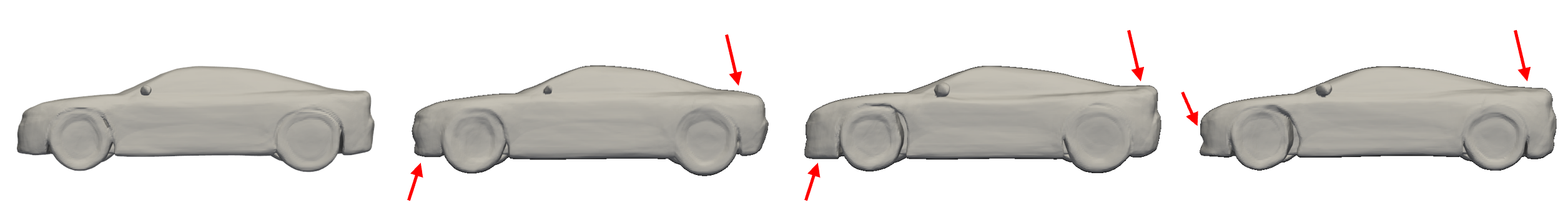}
  \caption{Results of optimization starting from four different initial shapes. The arrow indicate the points of change from the leftmost image. Figure \ref{fig: Anather results of optimization starting from four different initial shapes. The arrow indicate  the points of change from the leftmost image} in Appendix \ref{appendix: VehicleSDF generation} show another examples.
}
  \label{fig: different initial}
\end{figure}

\paragraph{Various 3D shape generation} Geometric parameters and 3D shapes have a one-to-many relationship, where various shapes can satisfy a single set of target parameters. VehicleSDF can generate various final shapes from different initialized latent vectors. Figure \ref{fig: different initial} shows the results generated from four different initial latent vector by inputting the target parameters from Table \ref{tab:vehicle_geometric}. It can be observed that the rear and front silhouettes differ locally. This demonstrates that VehicleSDF can output 3D shapes that satisfy the geometric parameters with variations.

\paragraph{Drag coefficient prediction} \label{Drag predictor} As the feature extractor for the drag coefficient prediction model shown in Figure 7, this research compares CLIP \cite{radford2021learning} and Vision Transformers (ViT) \cite{dosovitskiy2020image}. CLIP model embeds images into 512-dimensional vectors, while ViT model embeds them into 1024-dimensional vectors. We then employed a learnable LightGBM model to predict \(C_d\) values from these vectors. For CLIP features, the test R$^2$ and \text{MSE} were 0.55 and 0.0026, respectively, and for ViT features they were 0.58 and 0.0024. Since ViT features achieve a slightly better performance, we use them as part of our drag coefficient prediction model in VehicleSDF.

\begin{figure}
  \centering
  \includegraphics[width=1.0\textwidth]{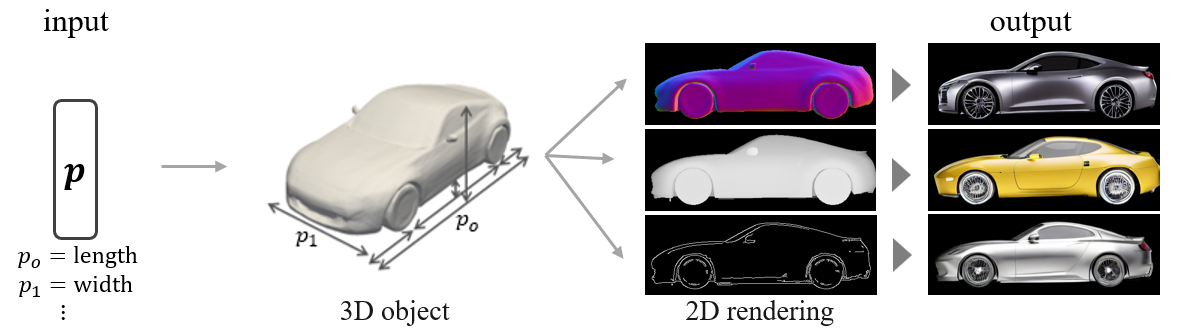}
  \caption{Generating styling images by ControlNet with the input of 3D shapes.}
  \label{fig: image generation}
\end{figure}

\paragraph{Image generation} Figure \ref{fig: image generation} shows the results of generating detailed styling images using ControlNet \cite{zhang2023adding} based on 2D renderings of the side views of 3D shapes generated by VehicleSDF, using normal map, depth map, and canny edge. These results demonstrate that the 3D shapes generated by VehicleSDF have sufficient quality for photo-realistic styling.

\section{Conclusion, Limitations and Future Work}
\label{Conclusion}
Generative AI tools that integrate engineering constraints, performance metrics, and aesthetics can accelerate the design process. This research demonstrates that we can modify current 3D shape generation techniques to incorporate engineering constraints. Additionally, we provided a proof of concept by generating \( C_d \) values and aesthetic vehicle renderings from the generated shapes. However, our approach does not currently allow direct optimization based on \( C_d \) values. Attempts to predict \( C_d \) from DeepSDF’s latent space, similar to geometric parameters, were not successful as the latent space could not capture the shape information necessary for drag coefficient prediction. Future improvements could leverage larger datasets like DrivAerNet++~\cite{elrefaie2024drivaernetlargescalemultimodalcar} and develop a fully differentiable pipeline from latent space to \( C_d \) prediction, which would be an exciting direction for future work.

%%%%%%%%%%%%%%%%%%%%%%%%%%%%%%%%%%%%%%%%%%%%%%%%%%%%%%%%
\newpage
\bibliographystyle{unsrt}
\bibliography{references}

%%%%%%%%%%%%%%%%%%%%%%%%%%%%%%%%%%%%%%%%%%%%%%%%%%%%%%%%
% \clearpage
\begin{appendices}
% \appendix

\section{DeepSDF Training Details}\label{appendix: DeepSDF}
DeepSDF considers a set of \(N\) shapes \(\{\bm{X}_i\}\), where each shape \(\bm{X}_i\) is sampled at \(K\) points \(\bm{x}_j \in \mathbb{R}^3\), and each point is assigned an SDF value \(s_j \in \mathbb{R}\). The SDF value \(s_j\) represents the signed distance from the surface, where points inside the surface have a negative sign, and points outside have a positive sign. The relationship between these \(K\) sampled points and the SDF values is expressed as: \(\bm{X}_i := \{(\bm{x}_j, s_j) : s_j = \text{SDF}^i(\bm{x}_j)\}.\) DeepSDF is trained by optimizing \(\theta\) and \(\bm{z}\) using the loss function:

\begin{equation}
\arg\min_{\theta, \bm{z}_i} \sum_{i=1}^{N} \left( \sum_{j=1}^{K} \mathcal{L}_\text{MSE}\left( f_\theta(\bm{z}_i, \bm{x}_j), s_j \right) + \frac{1}{\sigma^2_L} \|z_i\|_2^2 \right).
\end{equation}

The first term in the equation is the loss function \(\mathcal{L}_\text{MSE}\) applied to the output of the model \(f_\theta: \mathbb{R}^m \times \mathbb{R}^3 \to \mathbb{R}\) and the truth \(s_j\), and the second term is given by the \(L_2\)-norm of the latent \(\bm{z}_i\) and \(\sigma_L \in \mathbb{R}\) as  regularization. Given a set of spatial points $\bm{x}_j$, a 3D shape is generated by evaluating \(f_{\hat{\theta}}(\hat{\bm{z}}, \bm{x}_j)\) and extracting the isosurface where SDF = 0 using techniques such as ray casting or Marching Cubes~\cite{lorensen1998marching}. Here, $\hat{\theta}$ and $\hat{\bm{z}}$ represent the optimized values of $\theta$ and $\bm{z}$, respectively.

\section{Automatic Parameter Extractor Details}\label{appendix: Extracting parameters}
\begin{figure}
    \centering
    \begin{subfigure}[b]{0.47\textwidth}
        \centering
        \includegraphics[valign=c,width=\textwidth]{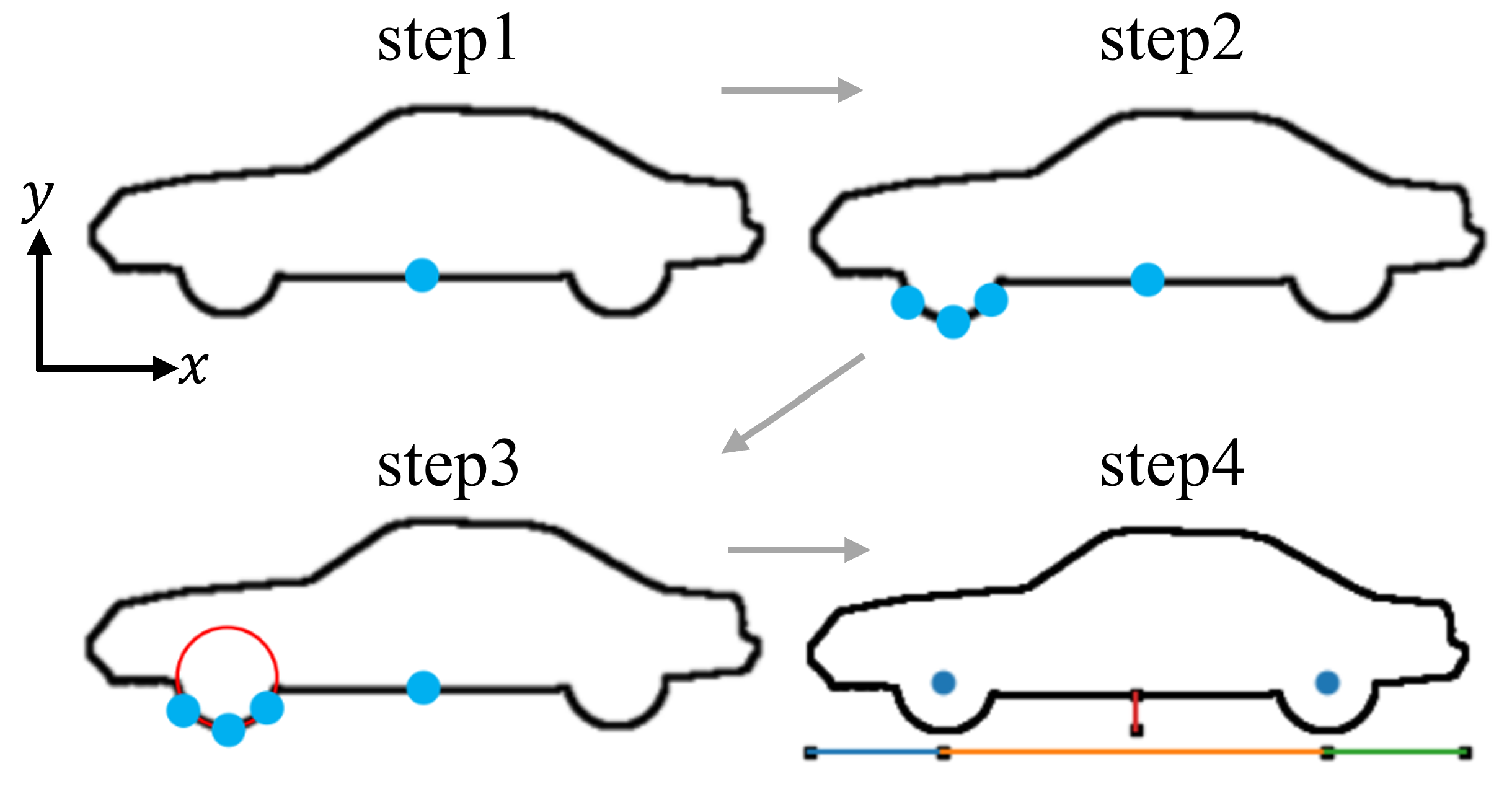}
        \caption{Automatic parameter extractor}
        \label{fig: Automatic parameter extractor}
    \end{subfigure}\;\;\;\;\;\;\begin{subfigure}[b]{0.36\textwidth}
        \centering
        \includegraphics[valign=c,width=\textwidth]{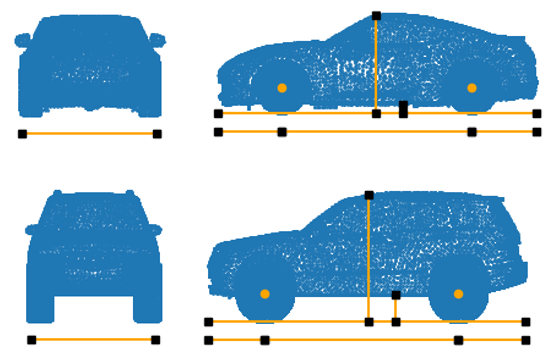}
        \caption{Examples of extracted parameters}
        \label{fig: Examples of extracted parameters}
    \end{subfigure}
    \caption{An example of extracting geometric parameters such as ground clearance, front and rear overhang, and wheelbase from ShapeNet shapes using method (a) and extraction method (b) is shown.}
    \label{fig: An example of extracting geometric parameters such as ground clearance, front and rear overhang, and wheelbase from ShapeNet shapes using method (a) and extraction method (b) is shown.}
\end{figure}

We used four steps to extract geometric parameters from the 3D geometry, as shown in Figure \ref{fig: Automatic parameter extractor}. 
\begin{enumerate}
    \item Extract the point in the vehicle with the central x-coordinate and the smallest y-coordinate.
    \item Extract 3 points whose y-coordinate is smaller than the point obtained in Step 1 and whose x-coordinate is smaller than the center of the vehicle. These are the three points on the front tire.
    \item Define a circle passing through all three points on the tire, and the center of the circle is the center of the front tire.
    \item Likewise, the center of the rear tire was determined, and the ground clearance, wheelbase, front overhang, and rear overhang were extracted.
\end{enumerate}
 On the other hand, length and height were extracted from the difference between the largest and smallest points on the x-axis and y-axis, respectively. Width was defined from the difference of the maximum and minimum points in the z-direction among the points with the same y-coordinate as the tire center to avoid mirrors. An example of automatic measurement using this algorithm is shown in Figure \ref{fig: Examples of extracted parameters}.

\section{Reconstruction and Interpolation Details}\label{appendix: reconstruction and interpolation} 
\begin{figure}[b]
  \centering
  \includegraphics[width=1.0\textwidth]{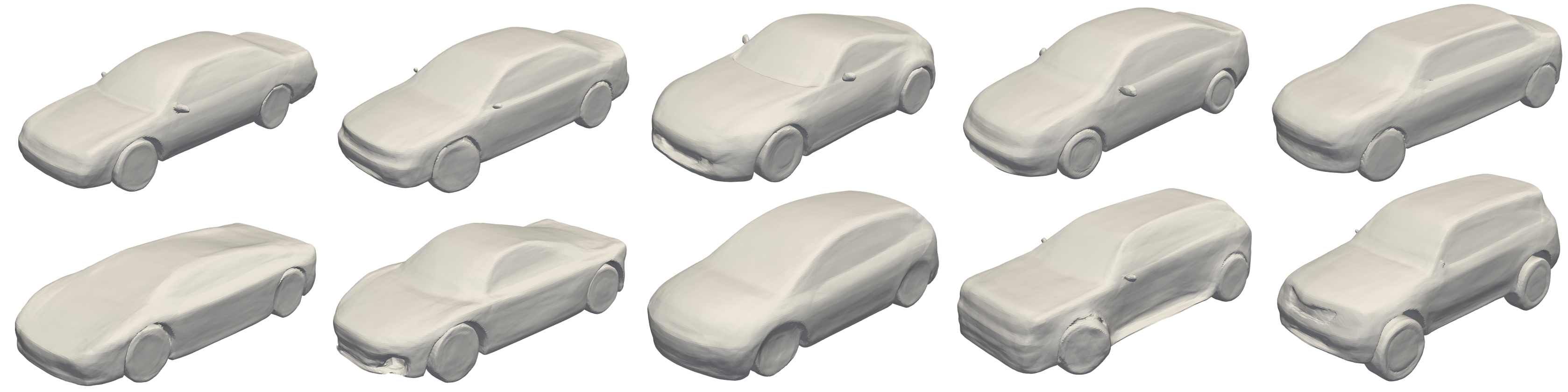}
  \caption{A illustration of diversity 3D shape generation by VehicleSDF.}
  \label{fig: A illustration of 3D shape generation by VehicleSDF with trained data.}
\end{figure}

\begin{figure}
  \centering
  \includegraphics[width=1.0\textwidth]{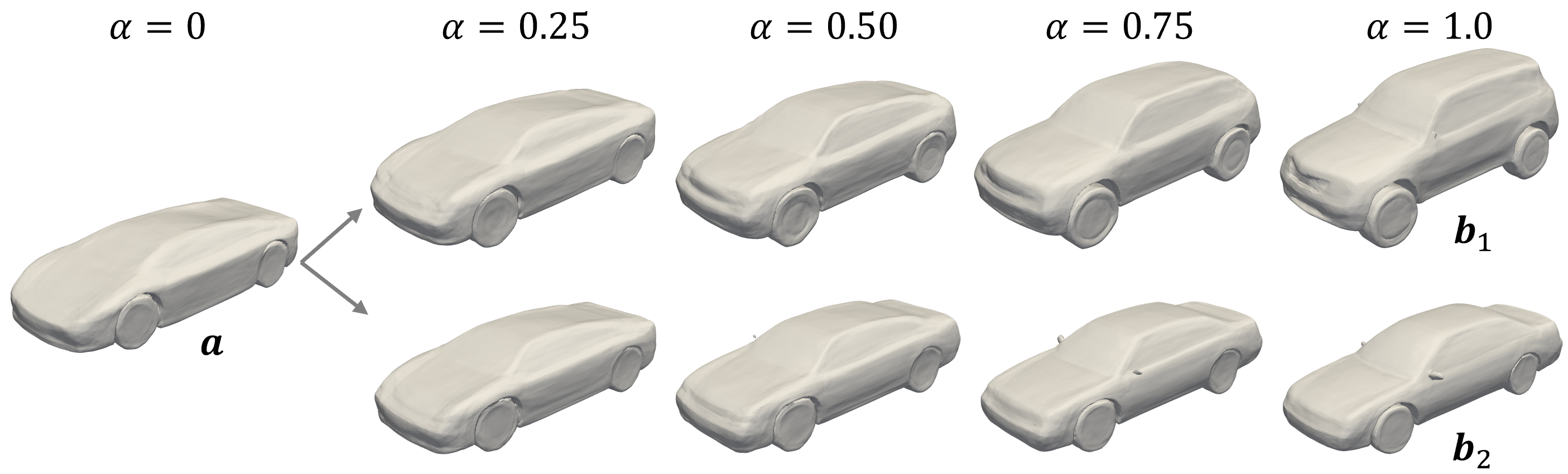}
  \caption{A illustration of interpolated generation from same initial shape to different target shapes}
  \label{fig: A illustration of interpolated generation from same initial shape to different shapes}
\end{figure}

The results of the 3d models generated from  VehicleSDF are shown in Figure \ref{fig: A illustration of 3D shape generation by VehicleSDF with trained data.}, showing that it is possible to generate diverse vehicle shapes. Figure \ref{fig: A illustration of interpolated generation from same initial shape to different shapes} shows the 3D models generated by interpolating between two designs. For interpolation, we used the interpolated latent \(\bm{r}\) defined as \( \bm{r} = \bm{a} + \alpha(\bm{b} - \bm{a})\), where \(\alpha \in (0, 1)\), \(\bm{a}\) and \(\bm{b}\) are optimized latent, respectively. We can see from this that the vehicle shape can be continuously changed in between two arbitrary shapes. We also augment the training dataset for training the parameter estimator model with interpolated shapes.

\section{Surrogate Parameter Estimator Training}\label{appendix: parame MSE}
\begin{figure}
  \centering
  \includegraphics[width=0.6\textwidth]{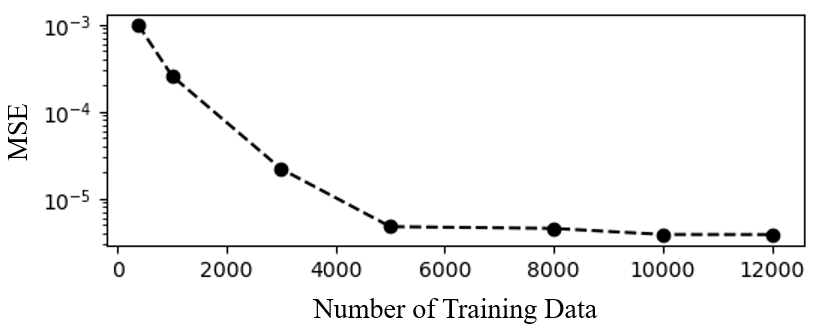}
  \caption{Shift in the parameter estimator's test MSE when augmenting dataset size from 374 to 12,000}
  \label{fig: Shift in MSE with Increasing Training Data}
\end{figure}

\begin{figure}
  \centering
  \includegraphics[width=1.0\textwidth]{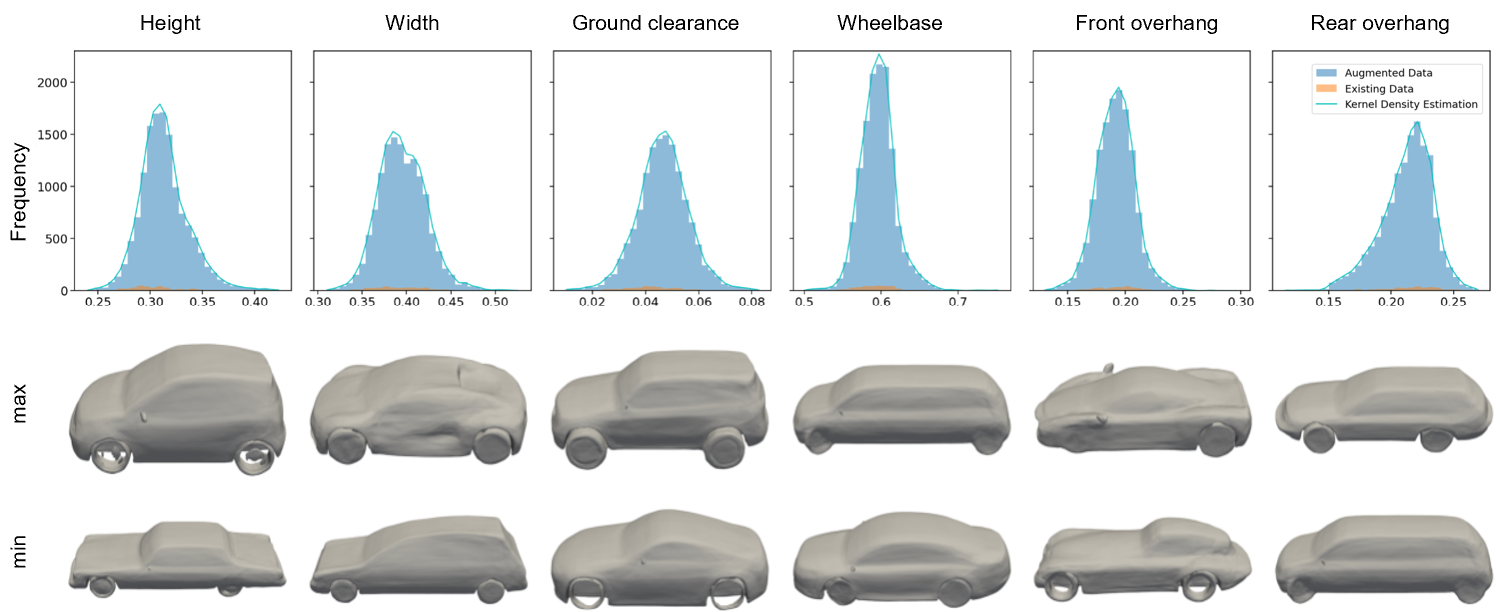}
  \caption{Distribution of the data used for the parameter estimation model. The top row shows the histograms of each parameter. The color differences indicate the existing data and the augmented data generated by the trained DeepSDF model. The middle row shows the 3D shape corresponding to the maximum value of each parameter, and the bottom row shows the 3D shape corresponding to the minimum value of each parameter.}
  \label{fig: Distribution}
\end{figure}

Figure \ref{fig: Shift in MSE with Increasing Training Data} shows the relationship between the MSE of the held-out test set, defined at 20\% of each data volume, with the amount of training data. The MSE saturated when the training data exceeded 10,000 examples. Figure \ref{fig: Distribution} show the distribution of each parameter in the augmented data, as well as the shapes with the maximum and minimum values for each parameter (except \emph{length} which is normalized to 1). The mean and variance of the augmented dataset \{length, height, width, ground clearance, wheelbase, front overhang, rear overhang\} were \{1.00, 0.31, 0.40, 0.05, 0.60, 0.19, 0.21\} and \{0.0, $5.4\times 10^{-4}$, $7.1\times 10^{-4}$, $9.0\times 10^{-5}$, $5.3\times 10^{-4}$, $2.7\times 10^{-4}$, $4.2\times 10^{-4}$\}, respectively.

\section{Validation of VehicleSDF}\label{appendix: VehicleSDF generation}
\begin{figure}
    \centering
    \begin{subfigure}[b]{0.46\textwidth}
        \centering
        \includegraphics[valign=c,width=\textwidth]{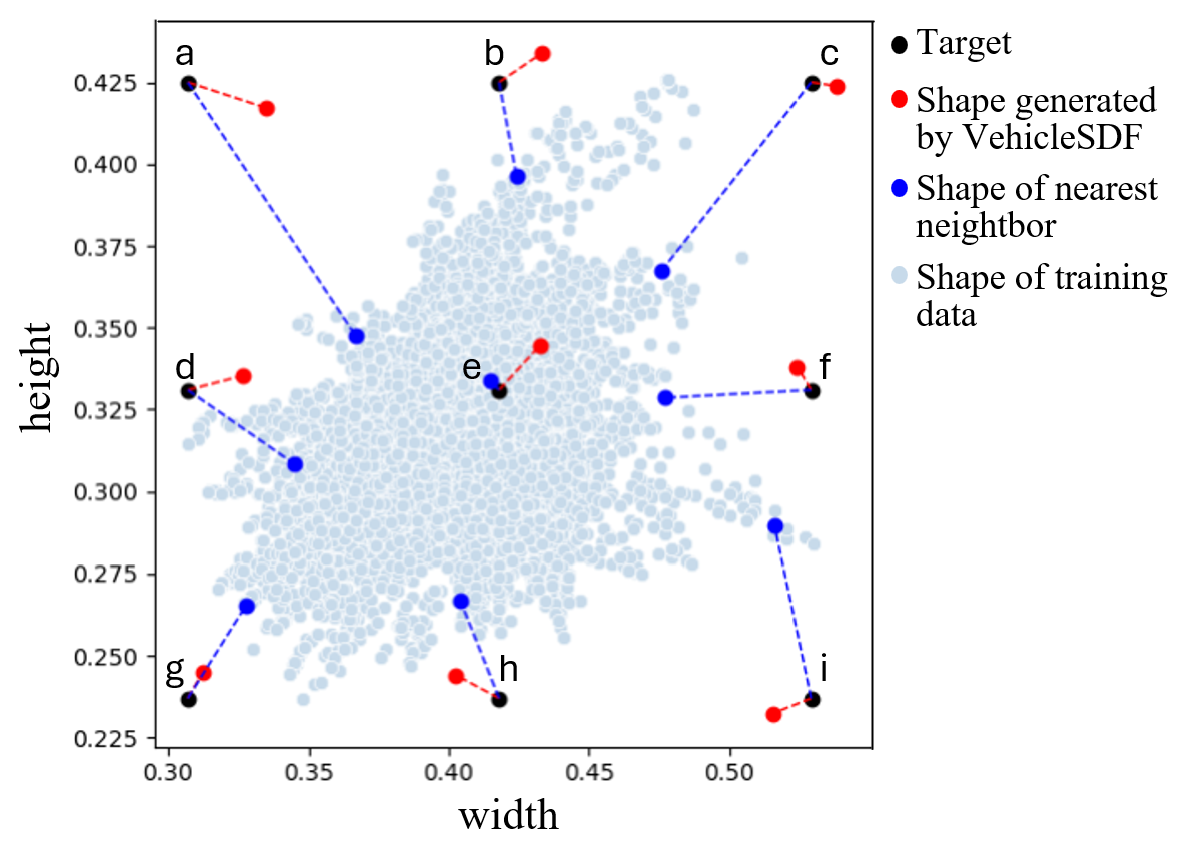}
        \caption{Target points and each shape parameters}
        \label{fig: target points and each shape parameters}
    \end{subfigure}\;\begin{subfigure}[b]{0.45\textwidth}
        \centering
        \includegraphics[valign=c,width=\textwidth]{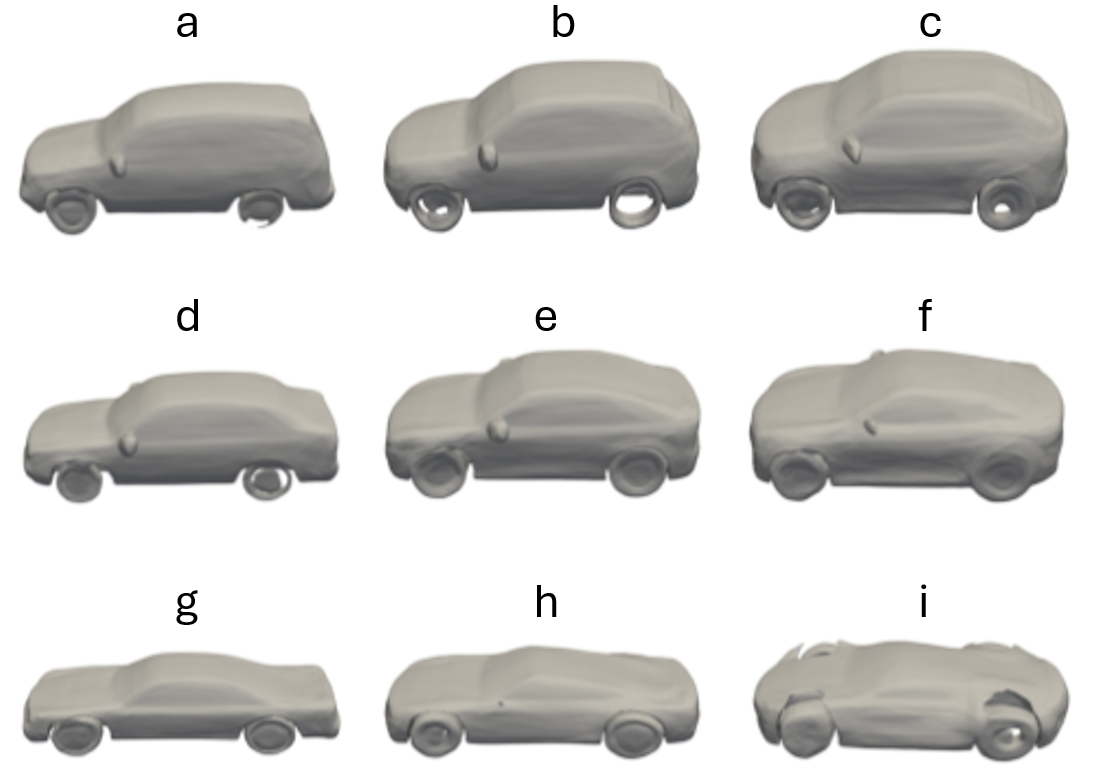}
        \caption{Shapes generated by VehicleSDF for each target}
        \label{fig: shapes generated by VehicleSDF for each target}
    \end{subfigure}
    \caption{3D shapes generated using VehicleSDF with nine target parameters are compared to nearest neighbor shapes in the training data to each target parameters (a). The nine target parameters consists of nine combinations of the maximum, minimum, and median values of height and width, plus the median values of the other parameters. The 3D shapes generated by VehicleSDF for each target is shown in (b).}
    \label{fig: nine points}
\end{figure}

Figure \ref{fig: nine points} shows the result of generating 3D shapes with target parameters that are within the distribution of the training data used for VehicleSDF as well as parameters that are outside of that distribution. The results show that it is possible to generate shapes for target parameters outside the distribution (other than target (e)). However, we observe that the shape of target (i) was unnatural due to insufficient training data. For targets within the distribution, a shape closer to the target can be found by directly searching for the nearest neighbor shape using Euclidean distance in parameter space, while for targets outside the distribution other than target f, the results generated by the VehicleSDF were closer to the target than the nearest neighbors.

\begin{figure}
  \centering
  \includegraphics[width=1.0\textwidth]{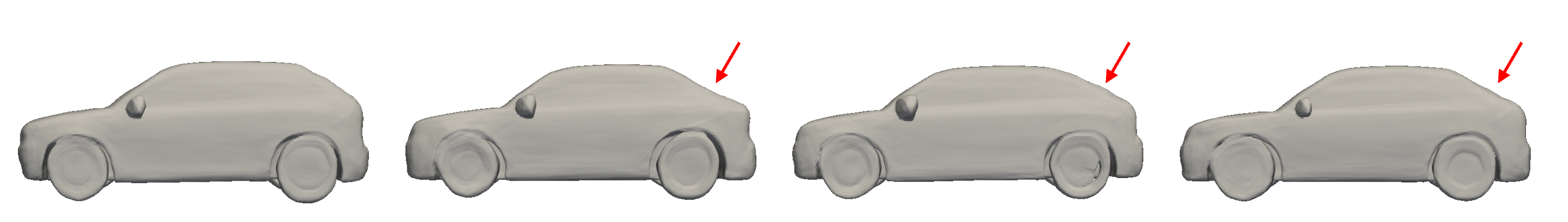}
  \caption{Another result for optimization starting from four different initial shapes. The arrow indicate the points of change from the leftmost image.}
  \label{fig: Anather results of optimization starting from four different initial shapes. The arrow indicate  the points of change from the leftmost image}
\end{figure}

\section{Drag Coefficient Predictor Training Details}\label{appendix: drag predictor}
\begin{figure}
    \centering
    \begin{subfigure}[b]{0.485\textwidth}
        \centering
        \includegraphics[valign=c,width=\textwidth]{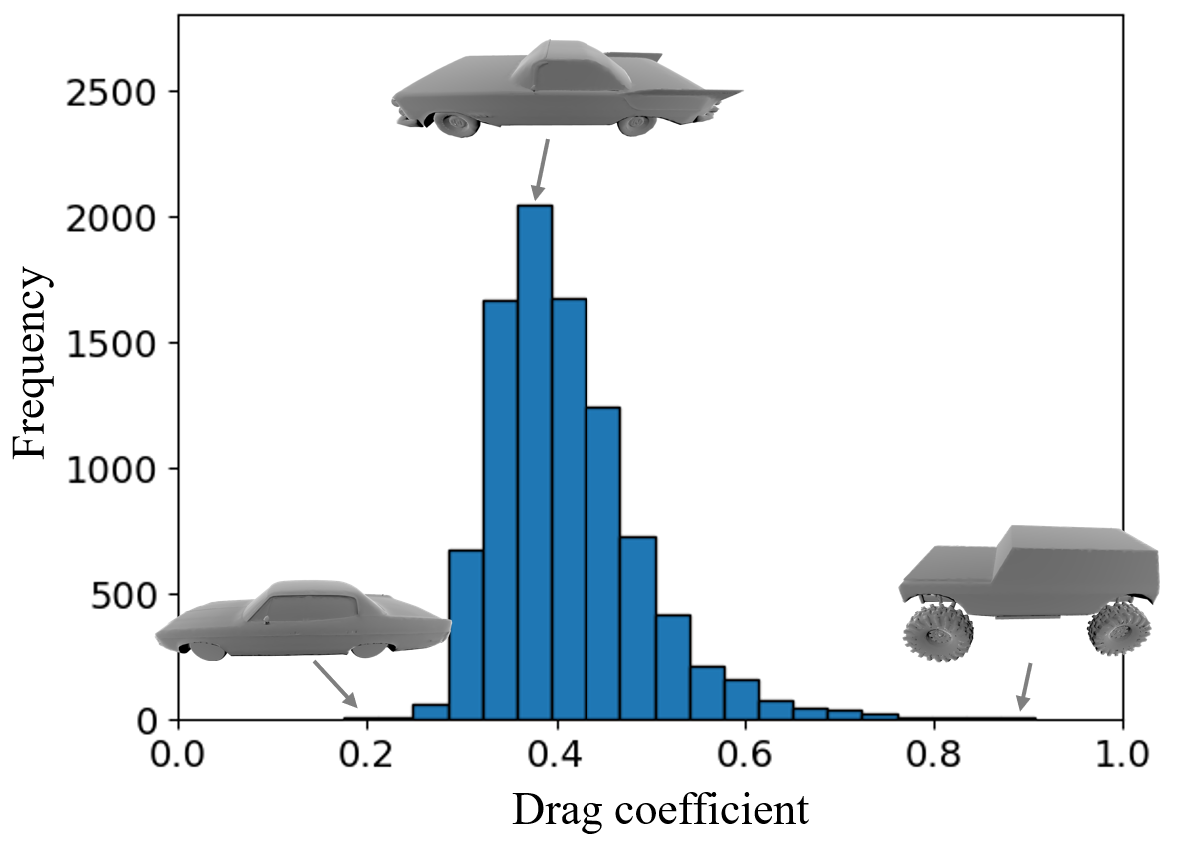}
        \caption{}
        \label{fig: a distribution of Cd}
    \end{subfigure}\;\begin{subfigure}[b]{0.46\textwidth}
        \centering
        \includegraphics[valign=c,width=\textwidth]{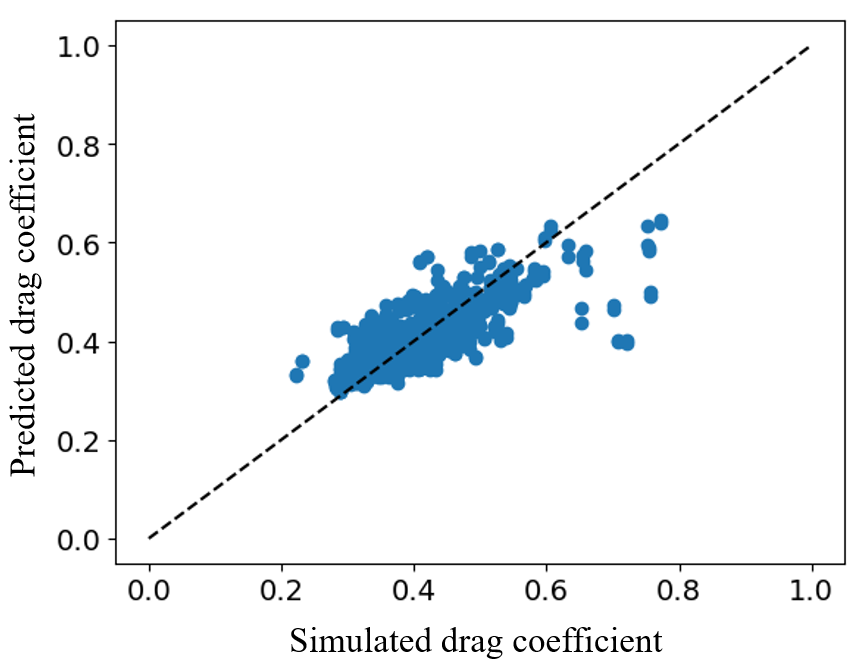}
        \caption{}
        \label{fig: The comparison between predicted and ground-truth}
    \end{subfigure}
    \caption{The drag coefficient ($C_d$) distribution for training drag predictor is shown in (a), and a comparison between predicted and ground-truth values of $C_d$ is shown in (b).}
    \label{fig: comparison}
\end{figure}

To train \(C_d\) value prediction model, we used a dataset of 9,070 shapes labeled with \(C_d\) values prepared in ~\cite{song2023surrogate}. This dataset was augmented by a factor of 4 using width increments (2x) and flips (2x), so that only a quarter of the shapes are completely unique. The distribution of \(C_d\) values for this dataset is shown in Figure \ref{fig: a distribution of Cd}. Also, we split the entire dataset into training, validation, and test sets following a ratio of 0.7:0.15:0.15 as in \cite{song2023surrogate} and trained the prediction model. A comparison of the prediction with the ground truth is shown in Figure \ref{fig: The comparison between predicted and ground-truth}.

%%%%%%%%%%%%%%%%%%%%%%%%%%%%%%%%%%%%%%%%%%%%%%%%%%%%%%%%
\end{appendices}
\end{document}